\documentclass[sigconf]{acmart}

\copyrightyear{2025}
\acmYear{2025}
\setcopyright{acmlicensed}\acmConference[ASPDAC '25]{30th Asia and South Pacific Design Automation Conference}{January 20--23, 2025}{Tokyo, Japan}
\acmBooktitle{30th Asia and South Pacific Design Automation Conference (ASPDAC '25), January 20--23, 2025, Tokyo, Japan}
\acmDOI{10.1145/3658617.3697668}
\acmISBN{979-8-4007-0635-6/25/01}
\usepackage{array}

\newcommand{\shk}[1]{\textcolor{black}{#1}}
\newcommand{\msb}[1]{\textcolor{black}{#1}}
\newcommand{\Msb}[1]{\textcolor{black}{#1}}
\newcommand{\fc}[1]{\textcolor{black}{#1}}
\begin{document}

% \title{Efficient Arbitrary Precision Acceleration for LLMs\\ on GPU Tensor Cores}
\title{Efficient Arbitrary Precision Acceleration for \\ Large Language Models on GPU Tensor Cores}

\author{Shaobo Ma$^1$, Chao Fang$^1$, Haikuo Shao$^1$, Zhongfeng Wang$^{1,2}$}
\affiliation{
\institution{
$^1$School of Electronic Science and Engineering, Nanjing University, Nanjing, China \\
$^2$School of Integrated Circuits, Sun Yat-sen University, Shenzhen, China}
\country{}
}
\email{
{shaoboma, fantasysee, hkshao}@smail.nju.edu.cn, zfwang@nju.edu.cn
}

\def \authors{Shaobo Ma, Chao Fang, Haikuo Shao, Zhongfeng Wang}
\renewcommand{\shortauthors}{Shaobo Ma, Chao Fang, Haikuo Shao, Zhongfeng Wang}

\begin{abstract}
\fc{Large language models (LLMs) have been widely applied but face challenges in efficient inference. While quantization methods reduce computational demands, ultra-low bit quantization with arbitrary precision is hindered by limited GPU Tensor Core support and inefficient memory management, leading to suboptimal acceleration.}
\fc{To address these challenges, we propose a comprehensive acceleration scheme for arbitrary precision LLMs. At its core, we introduce a novel bipolar-INT data format that facilitates parallel computing and supports symmetric quantization, effectively reducing data redundancy. Building on this, we implement an arbitrary precision matrix multiplication scheme that decomposes and recovers matrices at the bit level, enabling flexible precision while maximizing GPU Tensor Core utilization. Furthermore, we develop an efficient matrix preprocessing method that optimizes data layout for subsequent computations. Finally, we design a data recovery-oriented memory management system that strategically utilizes fast shared memory, significantly enhancing kernel execution speed and minimizing memory access latency.}
% \fc{Experimental results demonstrate that our scheme achieves a 200$\times$ speedup in 4k matrix multiplication with 1-bit weights and 2-bit activations compared to FP32. When applied to LLaMA2-7B, our scheme accelerates the 1-bit weight and 2-bit activation quantized model by 4$\times$ compared to PyTorch half-precision inference. This research provides a promising solution for efficient LLM deployment, potentially enabling wider applications of quantized LLMs.}
\fc{\Msb{Experimental results demonstrate our approach's effectiveness, with up to 2.4$\times$ speedup in matrix multiplication compared to NVIDIA's CUTLASS. }When integrated into LLMs, we achieve up to 6.7$\times$ inference acceleration. These improvements significantly enhance LLM inference efficiency, enabling broader and more responsive applications of LLMs.}
\end{abstract}


\maketitle

\section{Introduction}
% 关于Intro部分写作的一些建议（by shaohaikuo）：
% (1) 首先需要明确的一点，目前题目和摘要中的采用的是Arbitrary Precision Acceleration，任意精度的设计。那么在本文中如果设计和实验的目标是以ultra-low bit，超低比特为核心，那么在题目和相关描述中是否就可以明确以ultra-low bit为主。
% 举例，例如 Efficient Ultra-Low Bit Acceleration for Large Language Models on GPU Tensor Cores
% (2) 另外一点，Transformers or LLMs，在全文介绍的时候需要统一，显然Transformers包含的概念应该是大一些的，如果我们做的实验主要focus在LLM上，需要确定以Transformers为主题是否合适。
% (3) 如果是以ultra-low bit为主题，那么还需要补充一些ultra-low bit在相关模型中的强应用背景说明。
% (4) Contributions需要整理更详细充实一些
% (5) 关于GPU tensor cores的问题，格式不支持、支持但低效、内存管理等方面的问题，相关介绍需要更易懂，有说服力一些，必要的时候可以引用一些文章
% 是否可以参考APNN提供一个全文工作内容的框图
% In recent years, various neural network models based on the \\ \textbf{Transformer\cite{attention}} architecture have rapidly developed, ranging from the previously popular BERT\cite{bert} family to the currently trending \textbf{Large Language Models (LLMs)} led by the GPT family\cite{gpt1,gpt2,gpt3,gpt3.5,gpt4}. 
\fc{In recent years, large language models (LLMs) based on the Transformer~\cite{attention} architecture have been rapidly developed and gained popularity, with the GPT family~\cite{gpt1,gpt2,gpt3,gpt3.5,gpt4} being one of the most prominent examples.}
\fc{These models have demonstrated remarkable performance across a wide range of natural language tasks.} 
\fc{However, the increasing size and complexity of LLMs pose significant challenges for efficient inference. To address this issue, various acceleration methods~\cite{distillation1,prune1,prune2,gptq,qlora,smoothquant} have been proposed, among which model quantization~\cite{gptq,qlora,smoothquant,squeezellm,omniquant,awq,huang2024precision} has emerged as a promising approach. By reducing the data bit-width, quantization explicitly decreases memory consumption and computation time while minimizing the impact on the overall model structure and performance.}
% To meet the growing demand for efficient inference of LLMs, previous research has proposed various acceleration methods from different perspectives.} 
% To met the demand for improving the inference performance of Transformer models, previous research has proposed various inference acceleration methods for LLMs from different perspectives.
% , including model distillation\cite{distillation1,distillation2}, model pruning\cite{prune1,prune2}, and model quantization\cite{gptq,qlora,smoothquant,squeezellm,tsld,onebit}. 
% Among these approaches, inference acceleration for LLMs based on model quantization has minimized the impact on the overall model structure while explicitly reducing memory consumption and computation time by decreasing data bit-width.

% Recently, the quantization of LLM models has reached ultra-low bits (such as 2-bit and 3-bit), while the main computational resource, GPUs, only supports limited data formats (such as 1-bit and 4-bit), resulting in many theoretically effective quantization methods being unable to match the corresponding data formats in practical deployment, thus failing to achieve the expected acceleration. Specifically, the following issues exist:

\fc{As quantization methods continue to evolve, the quantization of LLM models has reached ultra-low bit-widths~\cite{omniquant, awq}, such as 2-bit and 3-bit.}
\fc{However, the mismatch between these advanced quantization methods and the capabilities of mainstream GPU devices leads to several challenges in practical deployment~\cite{qserve,bstc,apnn-tc}, which manifests in the following aspects:}
\fc{\textbf{1) Limited data format support in GPU Tensor Cores (TCs).} TCs are the key hardware components in modern GPUs for accelerating matrix multiplications (MatMuls), which dominate the operations in LLMs. Starting with the Turing architecture, Nvidia GPU TCs support some low-precision tensor operations, such as INT1 and INT4, and even bit-level operations~\cite{turing}. However, TCs still lack support for certain data formats that are widely used by ultra-low-bit quantized LLMs, such as INT2~\cite{tsld} and INT3~\cite{gptq}. When deploying these models on GPUs, the low-bit quantized data needs to be converted into higher-bit data formats supported by TCs for computation. This data format conversion introduces additional computational overhead, preventing some ultra-low-bit quantized LLMs from achieving optimal inference acceleration on GPU platforms.}

\fc{\textbf{2) Inefficient GPU memory management schemes.} When deploying LLMs on GPUs, performance bottlenecks arise not only from MatMuls but also from data access. GPUs employ a multi-level memory hierarchy~\cite{gpu_hierarchy}, with different levels varying in capacity and bandwidth. Effective memory management significantly impacts data access latency. 
% For example, on the NVIDIA GTX780 GPU, the memory-bound G-BLASTN achieves an overall 14.8x speedup compared to the sequential NCBI-BLAST by coordinating the use of GPU texture and shared memory~\cite{gblas}. 
For example, by carefully coordinating the use of different GPU memory levels, such as texture and shared memory, memory-bound applications can achieve significant speedup up to 14.8$\times$, compared to unoptimized implementations~\cite{gblas}.
Consequently, some optimization schemes that focus solely on MatMuls while neglecting the importance of GPU memory management fail to achieve the desired performance improvements in practice and may even result in slower execution times than the original, unoptimized one~\cite{pommdnn}.}

\msb{To address these challenges, we first propose a novel data format called bipolar-INT, which features a symmetric range and eliminates redundant sign bits, thereby reducing redundancy and facilitating parallel computing. Secondly, to overcome the limited precision support of GPU TCs, we achieve arbitrary precision MatMul through bit-level decomposition and recovery of matrices, saving memory while maintaining flexibility. Next, we introduce an efficient matrix preprocessing method that preprocesses input matrices by decomposing and reassembling them in advance, reducing communication costs and facilitating subsequent computations. Finally, we propose a memory management approach oriented by efficient data recovery, aiming to maximize the use of faster shared memory for tasks originally handled by global memory, greatly enhancing kernel execution speed. \Msb{Ultimately, compared to NVIDIA's MatMul acceleration design CUTLASS, our scheme achieves up to a 2.4$\times$ }speedup in MatMul and up to a 6.7$\times$ inference acceleration when integrated into LLMs.}

In summary, the main contributions of this paper are as follows:
\begin{itemize}
    \item We propose a novel data format called bipolar-INT \fc{for efficient arbitrary precision MatMuls}, featuring a symmetric range and eliminating redundant sign bits. (Section 3.1)
    % \item \msb{We propose a design for arbitrary precision MatMul on GPU}, saving memory while maintaining flexibility. (Section 3.2)
    \item \fc{We present an innovative design for arbitrary precision MatMuls on GPUs, optimizing memory usage without sacrificing flexibility. (Section 3.2)}
    \item \shk{An efficient matrix preprocessing method is designed to reduce communication costs and facilitate subsequent matrix computations. (Section 4.1)}
    \item \shk{A hierarchical memory management strategy oriented by efficient data recovery is proposed, which greatly enhances kernel execution speed. (Section 4.2)}
\end{itemize}
\section{BACKGROUND}
\subsection{Ultra-Low Bit Quantized LLMs}
% The enormous number of parameters in large language models (LLMs) leads to extremely high inference costs. Model quantization has shown considerable advantages in accelerating these models and has been extensively researched. 
% \shk{
% Model quantization has shown significant benefits in reducing computational complexity and memory footprint for accelerating LMMs, which has been extensively researched. 
% These quantization works are primarily based on two approaches: quantization-aware training (QAT)\cite{llm-qat,onebit} and post-training quantization (PTQ)\cite{gptq,qlora}.
% }
\fc{The rapid development of LLMs has led to a significant increase in model size and computational complexity, posing challenges for efficient inference. To address this issue, model quantization~\cite{llm-qat,onebit,gptq,qlora} has emerged as a promising approach, aiming to reduce the computational complexity and memory footprint of LLMs.}

\fc{Early LLM quantization methods focused on traditional techniques such as FP16 optimization~\cite{gpt3.5} and INT8 quantization. For example, GPT3.INT8()\cite{int8} addressed the issue of outliers in LLMs' quantization by employing mixed-precision computation. However, to further push the performance frontier, more aggressive quantization methods with lower bit-widths have been proposed, such as QLoRA\cite{qlora} with 4-bit, GPTQ~\cite{gptq} with 3-4 bit, SqueezeLLM~\cite{squeezellm} with 3-bit, TSLD~\cite{tsld} with ternary, and OneBit~\cite{onebit} with binary quantization for LLMs.}
\fc{Despite the promising results achieved by these ultra-low bit quantization methods, their optimal inference performance on GPUs has been hindered by the lack of suitable data formats supported by GPU hardware. This limitation calls for efficient acceleration designs that can bridge the gap between the quantization methods and the available GPU hardware capabilities, enabling LLMs to fully benefit from the reduced precision and computational complexity offered by ultra-low bit quantization.}

\subsection{GPU Hierarchy and Tensor Core}
% Graphics Processing Units (GPUs) have evolved from graphics rendering tools to essential infrastructures for complex computations in fields like artificial intelligence (AI). Their highly parallel structure makes them well-suited for simultaneously processing the extensive parallelized computations required in deep learning. GPUs feature a multi-level memory hierarchy, including global memory, shared memory, registers, and caches (L1 and L2). Global memory, the largest and slowest, is accessible by all threads, while shared memory, though smaller, is faster and accessible within a block, reducing latency. Registers, the fastest memory component, store frequently accessed variables for individual threads. L1 and L2 caches further expedite data access, with L1 being faster and L2 larger and shared among cores.
\fc{GPUs have become essential for AI workloads due to their highly parallel computing structure. Modern GPUs feature a multi-level memory hierarchy, including global memory, shared memory, registers, and caches (L1 and L2), each with different sizes and access speeds.}
Global memory, the largest and slowest, is accessible by all threads, while shared memory, though smaller, is faster and accessible within a block, reducing latency. Registers, the fastest memory component, store frequently accessed variables for individual threads. L1 and L2 caches further expedite data access, with L1 being faster and L2 larger and shared among cores~\cite{turing_archi,ampere_archi,hopper_archi}.

\fc{To accelerate deep learning workloads, NVIDIA introduced TCs in their GPUs. Optimized for MatMuls, a critical operation in LLMs, TCs leverage the massive parallelism of GPUs to significantly improve computational efficiency and inference performance~\cite{turing_archi,hopper_archi,ampere_archi,tc_1,tcgnn}.}
\fc{However, while TCs support low-precision data formats like INT1, INT4, and INT8, they lack support for efficient arbitrary precision operations. This limitation hinders the efficient acceleration of ultra-low bit quantized LLMs, necessitating a novel acceleration scheme that fully utilizes TCs to perform quantized LLMs with arbitrary precision~\cite{apnn-tc}.}
% NVIDIA's \msb{TC}s are specialized units within the GPU that significantly enhance deep learning performance. Optimized for matrix multiplication, \msb{TC}s accelerate critical operations in neural network training and inference. They can perform multiple operations per clock cycle, further boosting the GPU's parallel processing power. To support quantized model inference, \msb{TC}s in the Turing architecture added support for additional data formats, including int1, int4, and int8. For int1 precision, \msb{TC}s in the Turing architecture introduced support for XOR logical operations. This advancement has greatly promoted the development of model quantization, although it still does not support arbitrary precision operations.

\subsection{Arbitrary Precision Acceleration Schemes}
% Arbitrary precision (lower than INT8) acceleration designs have been extensively studied \cite{binaryconnect,o3bnn,deep-compression,bstc,turing,olacc,haq,lq-nets,dorefa,apnn-tc,bebert} to optimize neural network performance and accuracy for diverse application requirements. 
\shk{
Arbitrary precision acceleration schemes for data formats lower than INT8 have been extensively studied~\cite{binaryconnect,o3bnn,deep-compression,bstc,turing,olacc,haq,lq-nets,dorefa,apnn-tc,bebert} to optimize inference performance and maintain model accuracy for diverse application requirements.
% In addition to the widely supported precisions on modern GPUs (e.g., INT1, INT4, and INT8), these designs often utilize more varied precisions such as INT2, INT3, and INT5. 
In addition to the commonly supported precisions on modern GPUs (e.g., INT1, INT4), these designs often incorporate a broader range of precisions such as INT2, INT3.
}
% \msb{Prominent examples of arbitrary precision acceleration designs include APNN-TC~\cite{apnn-tc} with arbitrary precision, HAQ~\cite{haq} with 1-8 bits, and BSTC~\cite{bstc} and BTC~\cite{turing} with 1-bit precision. They designed MatMul kernels supporting more precision and integrated kernels into quantized neural networks to enhance the performance. However, previous work not only failed to achieve support for arbitrary precision (e.g., APNN-TC does not support W3A4), but also demonstrated suboptimal performance for large matrix parameters. This is because they did not use the most suitable data format and did not efficiently manage GPU memory. Based on this, this work first aims to find a more suitable data format for arbitrary precision MatMul, then truly implement kernel support for arbitrary precision, and finally further optimize the MatMul performance from the perspective of memory scheduling.}
\fc{Among these schemes, APNN-TC~\cite{apnn-tc} supports arbitrary precision, HAQ~\cite{haq} employs 1-8 bits, while BSTC~\cite{bstc} and BTC~\cite{turing} focus on 1-bit precision. These works designed MatMul kernels supporting various precisions and integrated them into quantized neural networks to enhance performance. However, they have limitations in terms of incomplete support for arbitrary precision, as exemplified by APNN-TC not supporting W3A4, and suboptimal performance for large matrix parameters due to unsuitable data formats and inefficient GPU memory management. To address these issues, our work aims to identify a more suitable data format for arbitrary precision MatMul, implement kernel support for true arbitrary precision, and further optimize MatMul performance through improved memory scheduling.}
% The aim of this paper is to be more flexible in terms of data formats than previous work while achieving faster model inference.
% \shk{
% Our work aims to provide a more flexible and efficient acceleration framework than previous works while achieving faster inference.
% }
\section{ARBITRARY PRECISION MATMUL}\label{APMM}
% The most time-consuming operation in Transformer model inference is matrix multiplication (MatMul). Additionally, the redundancy caused by unsupported data formats is primarily reflected in matrix multiplication. Therefore, the key objective of this paper is to implement arbitrary precision matrix multiplication during model inference. 

% In this section, we implement MatMul for arbitrary precision INT data using TCs. First, we decompose the input data bit by bit, breaking down arbitrary precision data into 1-bit data which is supported by TCs for MatMul, and design the corresponding data flow abstraction. Next, we introduce a bipolar-INT data format to replace the original INT format for MatMul and analyze the advantages of this approach.

\shk{
This section presents the arbitrary precision integers MatMul framework implemented on TCs.
First, we introduce an efficient data format, namely bipolar-INT, and demonstrate its advantages over traditional signed and unsigned integers for TCs' deployment.
Next, we propose a bit-wise MatMul decomposition method to separate the operands bit by bit based on our bipolar-INT, utilizing the bit-wise computing kernel supported by TCs, and further develop the data recovery dataflow.
}

\subsection{Bipolar-INT Data Format}
% We propose a novel \msb{data format called bipolar-INT, which is suitable for symmetric quantization and conducive to parallel computation,} for data quantization and \msb{MatMul}. Specifically, in typical INT data, each bit takes a value of 0 or 1, while in bipolar-INT data, a "0" is considered as "-1" during computation, meaning each bit can take on values of -1 or 1. 
% For a 2-bit example, the comparison of different data formats is shown in the TABLE \ref{tab:my_label} below. It is evident that bipolar-INT can represent odd numbers within a symmetric range. 

% \shk{
% We propose a novel and efficient data format called bipolar-INT, which is suitable for symmetric quantization and conducive to TCs' parallel computation. In traditional integers, each bit except the sign bit is valued as 0 or 1, whereas in the bipolar-INT, the "0" is interpreted as "-1" in calculations, allowing each bit to be either -1 or 1. Specifically, for a n-bit bipolar-INT data $x=\{x^{(n-1)}, \dots, x^{(1)}, x^{(0)} \}$, its decimal value can be obtained by}
% \vspace{-1em}

\fc{We introduce a novel and efficient data format called bipolar-INT for arbitrary precision MatMul computations. Compared to original signed and unsigned integers, bipolar-INT is more suitable for \msb{LLM quantization and} parallel computing due to its symmetric \msb{range} and unified operations, making it particularly advantageous for deployment on TCs.}

\begin{figure}[t]
    \centering
    \includegraphics[width=0.90\linewidth]{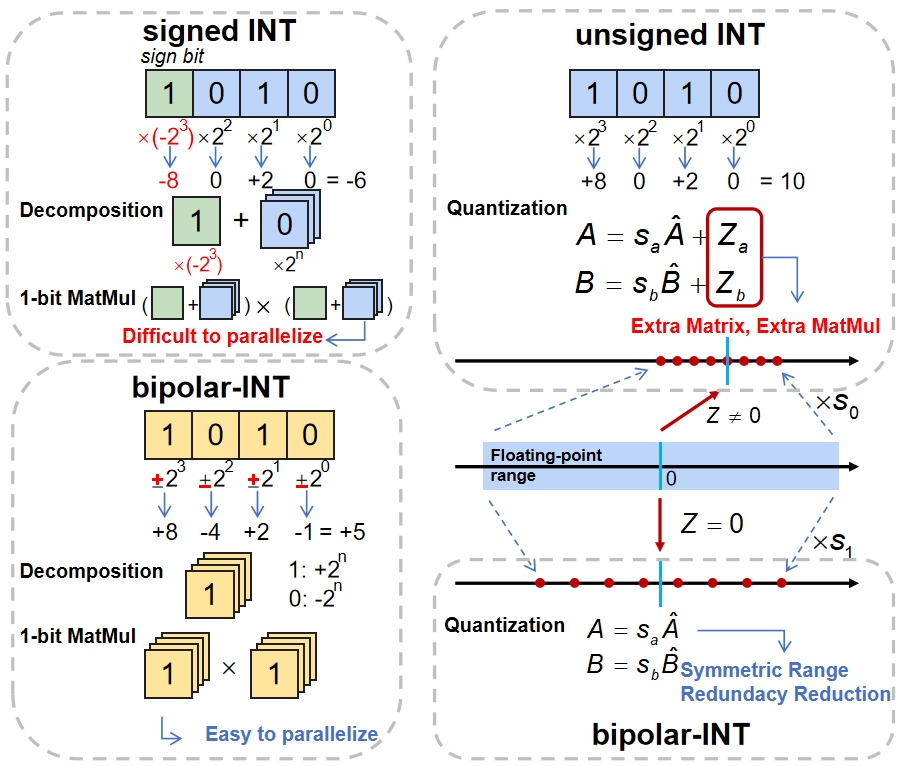}
    % \caption{Compared to original signed and unsigned integers, bipolar-INT is suitable for parallel computing due to symmetric quantization and unified operations.}
    \caption{\shk{Comparison between bipolar-INT and traditional integers. Bipolar-INT is well-suited for TCs’ parallel computing due to its symmetric quantization and unified operations.}}
    \label{fig:bipolar compare}
\end{figure}

\fc{As shown in Fig.~\ref{fig:bipolar compare}, the key difference between bipolar-INT and traditional integers lies in the interpretation of each bit. In traditional integers, each bit except the sign bit is valued as 0 or 1, whereas in bipolar-INT, the "0" is interpreted as "-1" in calculations, allowing each bit to be either -1 or 1. Specifically, for an n-bit bipolar-INT data $x={x^{(n-1)}, \dots, x^{(1)}, x^{(0)} }$, its decimal value can be obtained by}
\begin{equation}
    {(x)}_{D}=\sum\nolimits_{i=0}^{n-1} (2x^{(i)}-1)\cdot 2^{i}.
\end{equation}

\fc{For signed INT quantization, as illustrated in Fig. \ref{fig:bipolar compare}, due to the use of two's complement arithmetic, the sign of the MSB matrix after decomposition is opposite to the signs of the other bits. This requires separate handling during MatMul and matrix reconstruction, which is highly unfavorable for the parallel computation of single-bit matrices at each bit. Similarly, for unsigned INT quantization, the presence of an additional zero-point offset introduces extra multiply-accumulate operations during MatMul, which is detrimental to the optimization of MatMul.}

% \fc{In contrast, the zero-centered symmetric representation of bipolar-INT eliminates the need for additional zero-point offset matrices and avoids the need for special sign bits and two's complement arithmetic. This leads to unified operations during MatMul, making bipolar-INT highly compatible with the quantization method of binary neural networks, where weights are often quantized into two values: (-1, 1). By quantizing the feature matrix into the bipolar-INT format, we can directly use standard bipolar-INT MatMul operations without introducing additional operations.}

% \msb{Specifically, for binary quantized neural networks, each weight is often quantized to two values (-1, 1), represented by (0, 1). If the feature matrix is quantized using traditional INT quantization, for example, compute the MatMul of $W=[-1,1]$ and $X=[1,0]$. The method proposed by APNN-TC\cite{apnn-tc} introduces an additional all-ones matrix $J=[1,1]$. When read as an int type, the matrix X is read as $\hat{W}=[0,1]$, which relates to the actual data by $W=2\hat{W}-J$. Thus, the MatMul becomes $WX=2\hat{W}X-JX$, which not only introduces an additional matrix 
% $J$ occupying memory, but also introduces an extra MatMul operation 
% $JX$. However, the binary quantization aligns perfectly with the 1-bit bipolar-INT format. Therefore, when using bipolar-INT for MatMuls in binary quantized neural networks, we only need to quantize the feature matrix using bipolar-INT as well, without introducing additional matrices or MatMuls.}

\shk{Moreover, for binary quantized neural networks, weights ($W$) are often quantized to values of either -1 and 1, which are encoded as 1-bit 0 and 1, respectively. If the feature matrix $X$ is quantized to (0,1) with its decimal value also being (0,1), then the MatMul of $W$ and $X$ will be inconsistent during computation. 
Conventional methods like APNN-TC\cite{apnn-tc} introduce an additional all-ones matrix $J=[1,1]$ to tackle this problem. The matrix $W$ is decoded as $\hat{W}=[0,1]$, which relates to its actual value by $W=2\hat{W}-J$. Thus, the MatMul becomes $WX=2\hat{W}X-JX$, which not only introduces an additional matrix $J$ occupying memory, but also introduces an extra MatMul operation $JX$. 
In contrast, the binary quantized $W$ aligns perfectly with our 1-bit bipolar-INT format. Therefore, when using bipolar-INT for MatMuls in binary quantized neural networks, the feature matrix $X$ is also quantized using bipolar-INT, without introducing additional matrices or MatMuls for precise computation.}
\subsection{Bit-Wise MatMul Reconstitution}\label{subsec:Decomposition}

\begin{figure}[th]
    \centering
    \includegraphics[width=0.90\linewidth]{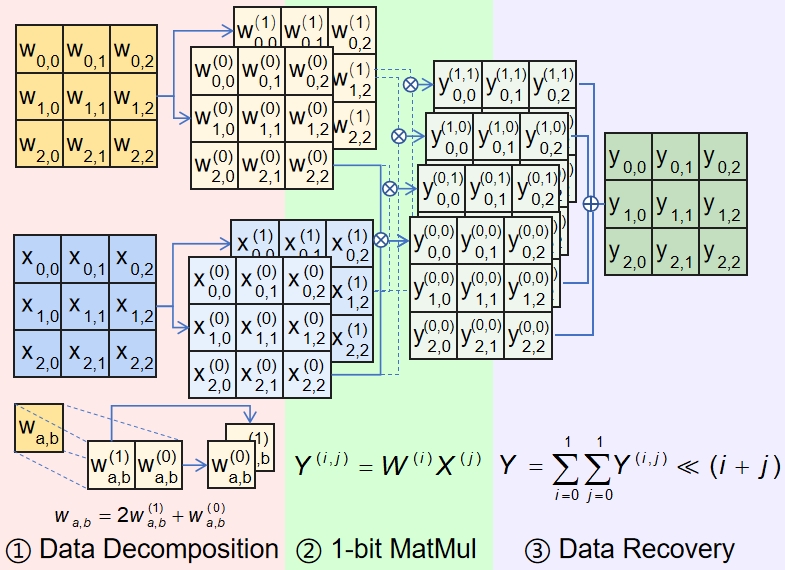}
    % \caption{Illustration of the computation process of arbitrary precision MatMul. Here matrices W and X are both 2-bit, which can be extended to arbitrary bit widths.}
    \caption{\shk{Illustration of the decomposition and recovery process for bipolar-INT MatMuls. Here matrices W and X are both 2-bit, which can be extended to arbitrary bit widths.}}
    \label{fig:AP-bit-matmul}
\end{figure}

\fc{A bit-wise MatMul reconstitution method is further proposed based on our bipolar-INT format. This method consists of three main steps: data decomposition, 1-bit MatMul, and data recovery.}
\fc{Fig. \ref{fig:AP-bit-matmul} illustrates the complete computation process of arbitrary precision MatMul using our proposed method. In this example, we consider a MatMul operation involving 2-bit matrices $W$ and $X$, ultimately calculating a 32-bit output $Y=WX$.}

\fc{The data decomposition step involves separating the operands bit by bit, enabling the utilization of the bit-wise computing kernel supported by TCs. Specifically, $W$ and $X$ are decomposed into two matrices each, denoted as $W^{(i)}$ and $X^{(j)}$, respectively.}
\fc{After the data decomposition, pairwise 1-bit MatMul operations are performed on the decomposed operands. Nvidia GPUs support the selection of either AND or XOR logic for 1-bit MatMul operations within TCs. By computing the pairwise MatMul of $W^{(i)}$ and $X^{(j)}$, we obtain a 32-bit intermediate result matrix $Y^{(i,j)}$ for each pair of bits.}
\fc{To reconstruct the final result from the intermediate bit-wise computations, a data recovery dataflow is employed. This step involves reconstructing the output matrix $Y$ from the intermediate results $Y^{(i,j)}$ by shifting each $Y^{(i,j)}$ based on its corresponding bit positions $(i,j)$ and then summing up all the shifted matrices.}

\fc{Although the example in Fig. \ref{fig:AP-bit-matmul} involves 2-bit matrices, the same principles can be extended to arbitrary bit widths. Our bit-wise MatMul decomposition method, based on the bipolar-INT format, effectively utilizes the bit-wise computing capabilities of TCs, providing a flexible and efficient solution for arbitrary precision MatMul operations.}

\section{GPU MEMORY SCHEDULING}
% The GPU is composed of a multi-level memory hierarchy, with different memory levels varying in size and processing speed. Global memory, accessible by all threads, has the largest size but the slowest processing speed. In contrast, shared memory is only accessible by threads within a block, has much smaller size but much faster processing speed compared to global memory. Specifically, to match the processing speed of Tensor Cores, the GPU includes a set of extremely fast registers called fragments, which are dedicated to storing the inputs, outputs, and intermediate results of Tensor Core operations.

This section focuses on how to efficiently transfer data between different levels of the memory hierarchy in the arbitrary precision \msb{MatMul} kernel to maximize processing speed. 
% The complete GPU multi-level memory scheduling scheme is illustrated in Fig. \ref{fig:GPU-Mem-Scheduling}. 

\subsection{Matrix Decomposition and Reassembly}

\fc{We present a matrix decomposition and reassembly strategy to efficiently transfer data between different levels of the memory hierarchy in the arbitrary precision MatMul kernel. This strategy aims to reduce memory access redundancy and maximize data transfer speed by preprocessing the original $n$-bit INT matrix.}

\begin{figure}[h]
    \centering
    \includegraphics[width=0.90\linewidth]{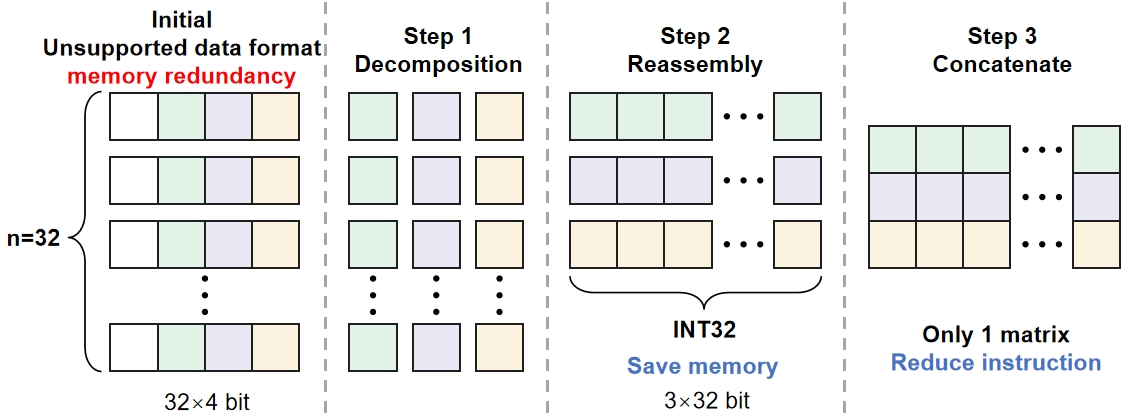}
    % \caption{Steps for matrix decomposition and reassembly to save memory and reduce instruction.}
    \caption{\shk{Procedure of matrix decomposition and reassembly to save GPU's memory and data transfer instructions.}}
    \label{fig:decomposition and reassembly}
\end{figure}

\fc{As shown in Fig. \ref{fig:decomposition and reassembly}, GPUs support more precisions than TCs when accessing memory, but they still do not support all possible precisions. For example, a 3-bit INT type does not have a suitable storage format and must be stored using a wider data format (such as 4-bit or 8-bit), introducing redundant memory access overhead. Moreover, when the amount of data to be accessed is large, not only the storage precision support but also the suitability of the corresponding data format for data transfer must be considered.}
% While CPUs and GPUs have native support for 32-bit data types, which includes benefits such as data alignment, memory access efficiency, and optimized processor instruction sets, 4-bit or 8-bit data types do not enjoy such support, affecting data transfer speed.

\fc{To address these issues, our strategy involves three steps, as illustrated in Fig. \ref{fig:decomposition and reassembly}. In Step 1, we perform 1-bit decomposition on the original matrix, breaking down each bit and regrouping them with corresponding bits from other data to form $n$ 1-bit matrices. This step circumvents the issue of unsuitable data formats and eliminates memory redundancy caused by the lack of appropriate data formats.
Next, in Step 2, we reassemble the decomposed data using 32-bit unsigned INTs. This step ensures that the input data aligns with the GPU's native support, thereby enhancing data transfer speed.
Finally, in Step 3, we sequentially concatenate the processed $n$ matrices into a single matrix. This step not only further conserves memory but also simplifies $n$ data transfer instructions into a single instruction. Although the amount of data transferred remains the same, this concatenation improves data transfer speed and saves storage space.
By following these steps, our matrix decomposition and reassembly strategy effectively reduces memory access redundancy and maximizes data transfer speed in the arbitrary precision MatMul kernel.}

\subsection{Recovery-Oriented Memory Scheduling}

\fc{A recovery-oriented memory scheduling strategy is proposed to optimize data transfer and memory access in the arbitrary precision MatMul kernel on GPUs by performing the matrix recovery process in shared memory or fragments, thereby reducing global memory access and computation, and further accelerating the kernel's computation speed.}

\begin{figure}[t]
    \centering
    \includegraphics[width=1\linewidth]{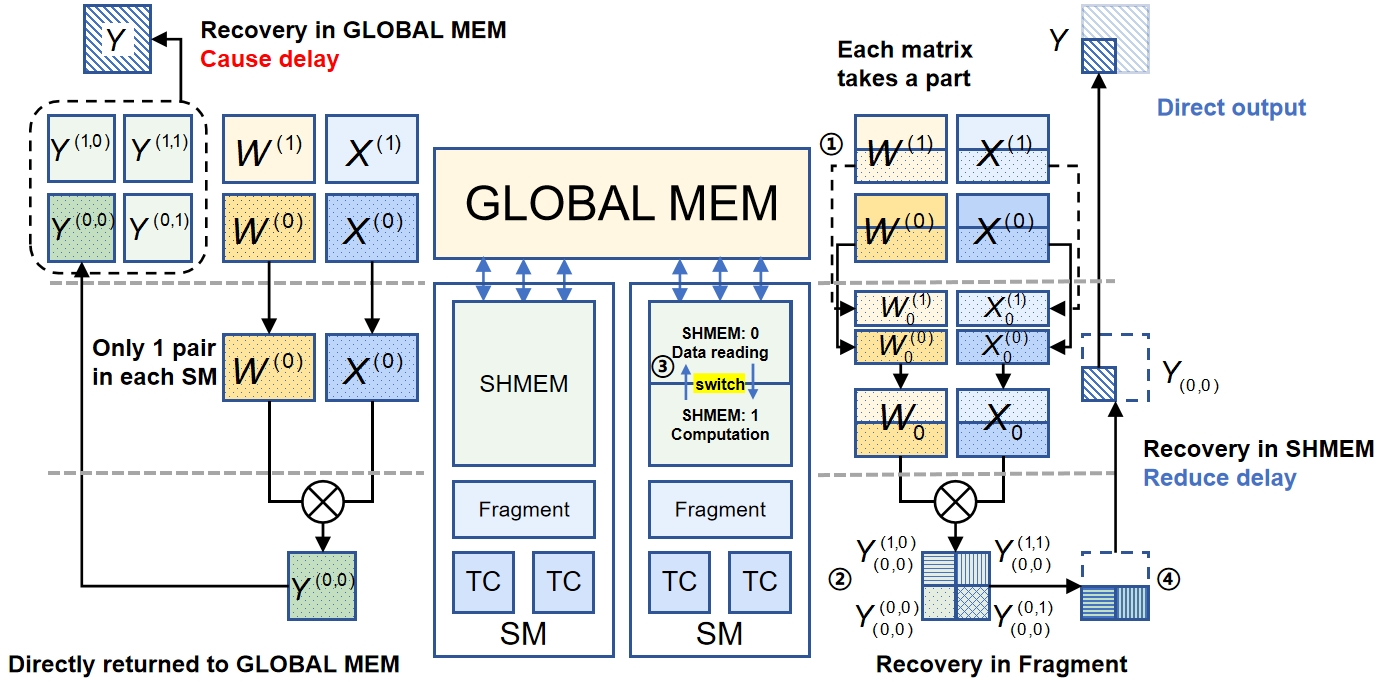}
    % \caption{Complete GPU multi-level memory scheduling design. Oriented by efficient data recovery, MatMul's latency is reduced.}
    \caption{\fc{Recovery-oriented memory scheduling strategy for arbitrary precision MatMul on GPUs, leveraging shared memory and fragments to reduce global memory access and accelerate computation.}}
    \label{fig:GPU-Mem-Scheduling}
\end{figure}

\fc{As shown in Fig. \ref{fig:GPU-Mem-Scheduling}, when implementing the arbitrary precision MatMul described in Sec. \ref{APMM} on a GPU, we perform 1-bit MatMul on the decomposed matrices to obtain intermediate result matrices. These intermediate result matrices are then shifted and summed to get the final result. In a naive approach, each streaming multiprocessor (SM) directly multiplies a pair of 1-bit matrices decomposed from the weights and features, and the MatMul result is directly returned to global memory for recovery. However, this design leads to each SM obtaining at most one intermediate result matrix, forcing the final matrix recovery to be performed in the slower global memory. This step introduces significant delays because accessing and processing data in global memory is much slower compared to shared memory.}

\fc{To address this issue, our recovery-oriented memory scheduling strategy aims to compute all intermediate result matrices within a single SM, as shown in \textcircled{1} of Fig. \ref{fig:GPU-Mem-Scheduling}. This requires each SM to compute all bitwise combinations of the weight and feature matrices. To efficiently manage shared memory, given its limited size, we divide the output matrix into blocks of size $b_m\times b_n$, with each SM responsible for computing the data within one block. If the number of blocks exceeds the number of SMs, the SMs are iteratively called to perform the computations.}

\fc{Due to the insufficient size of shared memory, the dimension $K$ needs to be partitioned. Each time, the SM only reads data from two matrices of size $n_{w,x}b_{m,n}\times b_k$, and the results of each computation are accumulated over $K/b_k$ iterations to produce the complete output value. In shared memory, the input weight and feature matrices of different bits are concatenated into two matrices and input into the Fragment to call the TC to perform 1-bit matrix multiplication. The resulting $n_wb_m\times n_xb_n$ matrix contains all the data needed to recover a $b_m\times b_n$ output block, as shown in \textcircled{2} of Fig. \ref{fig:GPU-Mem-Scheduling}. By sending these data back to shared memory for data recovery, we can obtain part of the final output directly, without involving global memory in the computation.}

\fc{To hide the data transfer latency from global memory to shared memory, we allocate two blocks of shared memory of the same size, as shown in \textcircled{3} of Fig. \ref{fig:GPU-Mem-Scheduling}. While one block is responsible for computation, the other block reads the next set of data. They then alternate in this manner, effectively overlapping data transfer and computation.}
\fc{Furthermore, to increase data reuse and reduce latency, we allow each Fragment to read the weight matrix of the same bit and all bits of the feature matrix, calculating all intermediate results corresponding to this bit of the weight matrix, as shown in \textcircled{4} of Fig. \ref{fig:GPU-Mem-Scheduling}. This way, we can perform the feature part of the data recovery in the Fragment, leaving the weight part of the recovery for shared memory computation.}

\fc{By employing this recovery-oriented memory scheduling strategy, we significantly reduce the memory access and computation in global memory, effectively leveraging the faster shared memory and fragments to accelerate the arbitrary precision MatMul kernel on GPUs.}

\section{EXPERIMENTAL RESULTS}

% In this section, we first evaluate the performance of the arbitrary precision \msb{MatMul} kernel under different precisions, highlighting its advantages in both performance and flexibility. Subsequently, we integrate this kernel into LLM and assess the inference performance of models at full precision, half precision, and various ultra-low bit quantizations, demonstrating the improvements in inference speed.
% For the evaluation, we chose the Nvidia RTX 3090 GPU, which utilizes the Ampere architecture. 
% For the large language model, we selected the full precision and half precision versions of the Llama2-7B, OPT-6.7B, BLOOM-7B model.
% \msb{In this section, we first evaluate the computational performance of our arbitrary precision MatMul design on matrices of different sizes and shapes under various precisions, verifying the effectiveness of our redundancy reduction and memory management design. Subsequently, we integrate this kernel into different large language models to assess its inference acceleration effect. The evaluation is conducted on an NVIDIA RTX 3090, which is in a ubuntu 18.04 system with Intel(R) Xeon(R) Silver 4210R CPU @ 2.40GHz, gcc-7.5.0, and using CUDA-11.8, CUTLASS-2.11.}
% \fc{We select the full-precision and half-precision versions of the Llama2-7B, OPT-6.7B, BLOOM-7B models as the LLM benchmark.}

\fc{In this section, we evaluate the performance of our arbitrary precision acceleration method for LLMs. Our experiments are conducted on an NVIDIA RTX 3090 GPU within an Ubuntu 18.04 system, using CUDA-11.8 and CUTLASS-2.11. 
% We employ both full-precision (FP32) and half-precision (FP16) versions of three representative LLMs: Llama2-7B, OPT-6.7B, and BLOOM-7B as our benchmark models. 
The evaluation is divided into two main parts: (1) an assessment of our arbitrary precision MatMul kernels, and (2) an analysis of its impact on LLM inference performance. Through these experiments, we aim to demonstrate the effectiveness of our approach in accelerating LLM computations across different precision levels.}
% , focusing on computational efficiency and performance gains.}

\subsection{Arbitrary Precision Kernel Evaluation}
% We compare the arbitrary precision MatMul designs proposed in this paper, such as W1A2 (1-bit weights, 2-bit activations), W2A2, W3A4, with FP32 and FP16 \msb{MatMul}s. Subsequently, we compare these designs with NVIDIA's Tensor Cores-accelerated ultra-low bit MatMul (CUTLASS INT1 and CUTLASS INT4) to demonstrate that the computational performance of our designs is superior when using the same \msb{TC}s. We then compare our designs with other \msb{TC}-based \msb{MatMul} acceleration designs such as APNN-TC~\cite{apnn-tc}, BSTC~\cite{bstc}, and BTC~\cite{turing}, highlighting the performance advantages of our work in similar applications. 
% Specifically, the APMM design in APNN-TC also supports various precisions like W1A2 and W2A2, but it does not support W3A4 computations, further proving the flexibility of our work.
\fc{In this subsection, we evaluate our arbitrary precision MatMul kernel design on both square and LLM-specific MatMul tasks, validating the effectiveness of our redundancy reduction and memory management techniques. \Msb{We compare our designs, including W1A2 (1-bit weights, 2-bit activations), and W2A2, with standard FP32 and FP16 MatMuls.} We then benchmark against NVIDIA's Tensor Cores-accelerated ultra-low bit MatMul (CUTLASS INT1 and CUTLASS INT4) to showcase our superior computational performance when using the same TCs. Finally, we compare with other TC-based MatMul acceleration techniques, such as APNN-TC~\cite{apnn-tc}, BSTC~\cite{bstc}, and BTC~\cite{turing}, to demonstrate the performance advantages of our work in low-bit arbitrary precision acceleration.}

\begin{table}[h]\scriptsize
    \centering
    \caption{Arbitrary precision kernel performance in comparison with FP and CUTLASS towards large square MatMuls.}
    % (Latency is the average of 1000 times. Speedup is based on FP32.)}
    \begin{tabular}{|c|p{0.7cm}<{\centering}|p{0.8cm}<{\centering}|p{0.7cm}<{\centering}|p{0.8cm}<{\centering}|p{0.7cm}<{\centering}|p{0.8cm}<{\centering}|}
        \hline
        \textbf{M/N/K} & \multicolumn{2}{|c|}{\textbf{1k/1k/1k}} & \multicolumn{2}{|c|}{\textbf{2k/2k/2k}} & \multicolumn{2}{|c|}{\textbf{4k/4k/4k}} \\
        \hline
        \textbf{Schemes} & \textbf{Latency} & \textbf{Speedup} & \textbf{Latency} & \textbf{Speedup} & \textbf{Latency} & \textbf{Speedup}\\
        \hline
        \textbf{FP32} & 121us & 1.00× & 779us & 1.00× & 5690us & 1.00×\\
        \hline
        \textbf{FP16} & 44.2us & 2.73× & 263us & 2.96× & 1960us & 2.90×\\
        \hline
        \textbf{CUTLASS INT4} & 15.8us & 7.61× & 66.5us & 11.7× & 386us & 14.7×\\
        \hline
        \textbf{CUTLASS INT1} & 9.3us & 13.0× & 36.9us & 21.1× & 161us & 35.3×\\
        \hline
        \hline
        % \textbf{W3A4 (ours)} & 12.4us & 9.74× & 50.4us & 15.4× & 184us & 31.0×\\
        % \hline
        \textbf{W2A2 (ours)} & 12us & 10.08× & 54.8us & 14.2× & 323us & 17.6×\\
        \hline
        \textbf{W1A2 (ours)} & 8.7us & 13.9× & 30.6us & 25.5× & 160us & 35.6×\\
        \hline
    \end{tabular}
    \label{tab:square MatMul}
\end{table}

\subsubsection{Square MatMul Performance}
% To demonstrate the computational performance of the \msb{MatMul} design proposed in this paper, we first tested its computation speed with input matrices that are all square matrices. 

% Table \ref{tab:square MatMul} presents the evaluation of our work against PyTorch floating-point MatMul and CUTLASS in computing large MatMuls. The Latency is the average time of 1000 computations, and the Speedup is benchmarked against FP32 MatMul under the same conditions.

% The configurations for input matrices M/N/K range from 128/128/128 to 4k/4k/4k. We compared the acceleration effects of various schemes relative to FP32 \msb{MatMul}, as shown in the Fig. \ref{fig:square}. In the figure, the results for FP32, FP16, and Nvidia's GEMM design (CUTLASS-gemm) are represented with linear statistical plots, while the designs from papers such as APNN-TC and our proposed design are represented with bar charts.
\fc{To demonstrate the computational performance of our proposed MatMul design, we first evaluated its speed on square matrices of various sizes. Table \ref{tab:square MatMul} presents the comparison of our work against PyTorch floating-point MatMul and CUTLASS for large MatMuls. The reported latency represents the mean execution time over 1000 iterations, and the speedup is calculated relative to FP32 MatMul under identical conditions.}

% \fc{Compared to FP32 and FP16, our MatMul design shows increasing speedup as matrix size grows. For 4k$\times$4k matrices, our W1A2 configuration achieves a remarkable 193$\times$ speedup over FP32 and about 70$\times$ over FP16. 
% When compared to CUTLASS, which is limited to int1 and int4 precisions due to TC constraints, our design demonstrates superior performance. For 4k$\times$4k matrices, our W1A2 configuration outperforms CUTLASS INT4 by more than 13$\times$. Notably, both our W1A2 and W2A2 configurations surpass CUTLASS INT1 in performance, despite not having a bit-width advantage. Our W1A2 configuration is 5.5$\times$ faster than CUTLASS INT1, while our W2A2 configuration is 3.5$\times$ faster. Even our W3A4 configuration, which has a significantly wider bit-width, approaches CUTLASS INT1 performance, achieving approximately 88\% of its speed.}
\msb{Compared to FP32 and FP16, our MatMul design shows increasing speedup as matrix size grows. \Msb{For 4k$\times$4k matrices, our W1A2 configuration achieves a remarkable 35.6$\times$ speedup over FP32 and about 13$\times$ over FP16.} 
When compared to CUTLASS, which is limited to int1 and int4 precisions due to TC constraints, our design demonstrates superior performance. \Msb{Notably, our W1A2 configurations surpass CUTLASS INT1 in performance, despite not having a bit-width advantage. For 4k$\times$4k matrices, our W1A2 configuration outperforms CUTLASS INT4 by more than 2.4$\times$, while our W2A2 configuration is 1.24$\times$ faster. Even our W2A2 configuration, which has a significantly wider bit-width, approaches CUTLASS INT1 performance, achieving approximately 50\% of its speed.}}
\begin{figure} [t] %注意，这里设置是关键
	\centering
	\includegraphics[width=\linewidth,scale=1.00]{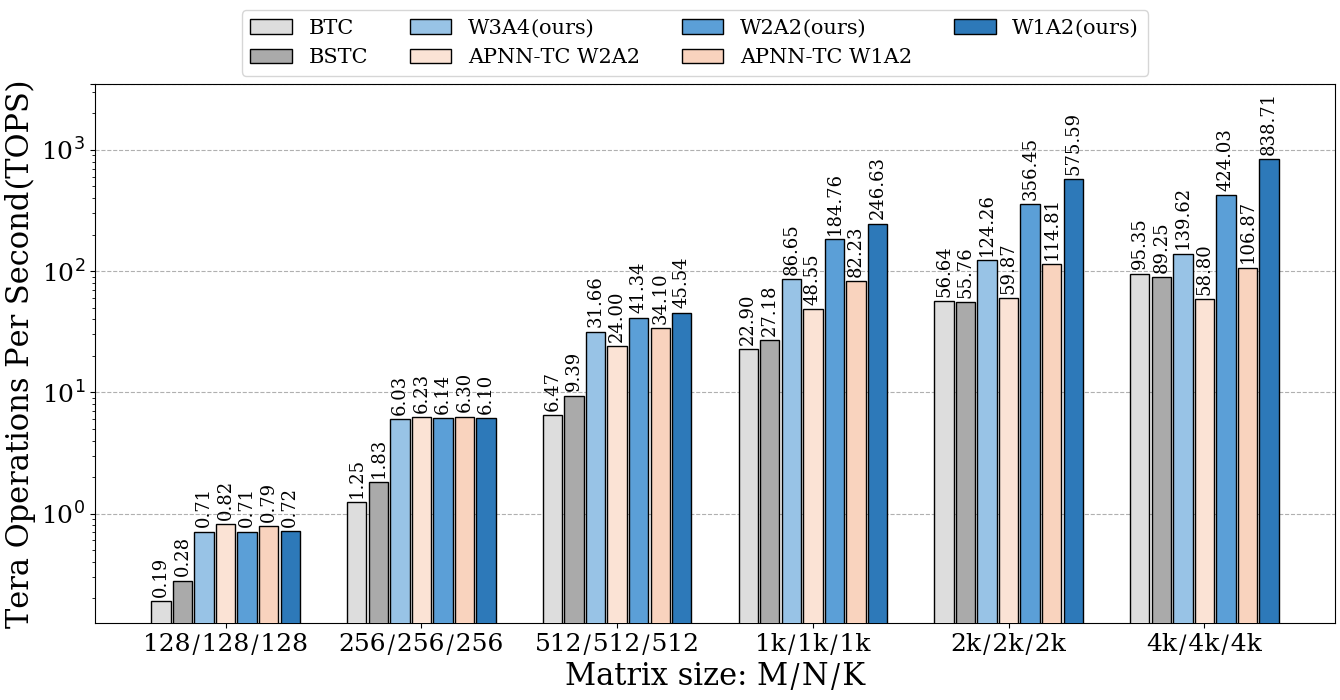}
	%[]里面的参数自己可根据需要调整
	% \caption{Comparison of our work with other approaches in the context of square MatMuls.}
        \caption{\shk{Comparison of throughput between our work and other methods in the context of square MatMuls.}}
	\label{fig:square}
\end{figure}

\fc{Fig. \ref{fig:square} illustrates the performance comparison of our work with other approaches on square matrices from 128$\times$128 to 4k$\times$4k, using Tera Operations Per Second (TOPS) as the metric. Note that APNN-TC is compared only with W1A2 and W2A2 configurations due to its limited precision support.}

\fc{While APNN-TC slightly outperforms our design for smaller matrices, our approach shows significant advantages for matrices 1k$\times$1k and larger, \Msb{with our W1A2 and W2A2 configurations achieving 7.9$\times$ and 7.2$\times$ speedups}, respectively. This is particularly relevant for LLMs, where large matrix operations are more common.
Furthermore, our design offers true arbitrary precision support, including configurations like W3A2 and W3A4, which APNN-TC cannot handle. This demonstrates not only superior performance but also greater flexibility, especially beneficial for the diverse computational needs of LLMs.}

\begin{table}[bth]\scriptsize\centering
    \centering
    \caption{Arbitrary precision kernel performance in comparison with FP and CUTLASS towards MatMuls in Llama2-7B.}
    % (Latency is the average of 1000 times. Speedup is based on FP32.)}
    \begin{tabular}{|c|p{0.7cm}<{\centering}|p{0.8cm}<{\centering}|p{0.7cm}<{\centering}|p{0.8cm}<{\centering}|p{0.7cm}<{\centering}|p{0.8cm}<{\centering}|}
        \hline
        \textbf{M/N/K} & \multicolumn{2}{|c|}{\textbf{1k/4k/4k}} & \multicolumn{2}{|c|}{\textbf{1k/10.5k/4k}} & \multicolumn{2}{|c|}{\textbf{1k/4k/10.5k}} \\
        \hline
        \textbf{Schemes} & \textbf{Latency} & \textbf{Speedup} & \textbf{Latency} & \textbf{Speedup} & \textbf{Lantency} & \textbf{Speedup}\\
        \hline
        \textbf{FP32} & 3.12ms & 1.00× & 8.21ms & 1.00× & 8.36ms & 1.00×\\
        \hline
        \textbf{FP16} & 1.07ms & 2.91× & 1.47ms & 5.58× & 1.58ms & 5.30×\\
        \hline
        \textbf{CUTLASS INT4} & 0.238ms & 13.1× & 0.574ms & 14.3× & 0.548ms & 15.3×\\
        \hline
        \textbf{CUTLASS INT1} & 0.097ms & 32.1× & 0.255ms & 32.2× & 0.188ms & 44.6×\\
        \hline
        \hline
        % \textbf{W3A4 (ours)} & 0.194ms & 16.1× & 0.523ms & 15.7× & 0.540ms & 15.5×\\
        % \hline
        \textbf{W2A2 (ours)} & 0.085ms & 36.9× & 0.222ms & 37.0× & 0.210ms & 39.8×\\
        \hline
        \textbf{W1A2 (ours)} & 0.051ms & 61.8× & 0.129ms & 63.6× & 0.109ms & 79.7×\\
        \hline
    \end{tabular}
    \label{tab:llm MatMul}
\end{table}

% \subsubsection{Matrix Multiplication from LLMs}
\subsubsection{LLM-specific MatMul Performance}
% Because most matrices in LLMs are not square matrices, to make our evaluation results more reliable, we split the Llama2-7B model layer by layer and extracted all its \msb{MatMul} parameters. We then evaluate the performance of the aforementioned \msb{MatMul} configurations based on these real matrix parameters found in LLMs. 
\fc{To more accurately reflect the computational demands of LLMs, we evaluate our MatMul designs on non-square matrices commonly found in these models. We extracted the MatMul parameters from each layer of the Llama2-7B model, providing a real-world benchmark for our evaluation.}

% Table \ref{tab:llm MatMul} shows the three sets of MatMuls in Llama2-7B that have the largest number of parameters and the greatest impact on inference time. 
% This paper evaluates the performance of this work with pytorch FP and CUTLASS MatMul under these configuration.
% It can be observed that, when the matrices are not square but instead are real matrices from the LLMs, the acceleration effect of our work slightly decreases. This is because \msb{TC}s are inherently more suited for handling square \msb{MatMul}s, as evidenced by the performance drop in CUTLASS INT1 and INT4. However, this does not diminish the outstanding performance of our work. Compared to CUTALSS, our work under the W1A2 and W2A2 configurations still surpasses CUTLASS INT1, which has a bit-width advantage. Additionally, under the W3A4 configuration, our performance also exceeds that of CUTLASS INT4. This improvement is due to our approach of splitting matrices to minimize data redundancy during computation, as well as implementing an efficient memory management scheme to further enhance kernel performance.
\fc{Table \ref{tab:llm MatMul} presents the performance comparison for the three most computationally intensive MatMul operations in Llama2-7B. These operations have the largest number of parameters and the greatest impact on inference time. The reported latency represents the mean execution time over 1000 iterations, and the speedup is calculated relative to FP32 MatMul under identical conditions.}

\fc{Our results show that while the acceleration effect slightly decreases for non-square matrices compared to square matrices, our approach still demonstrates outstanding performance. \Msb{For instance, our W1A2 configuration achieves speedups of 61.8$\times$, 63.6$\times$, and 79.7$\times$ for the three matrix sizes tested, significantly outperforming both FP16 and CUTLASS implementations.}}
\fc{Notably, our W1A2 and W2A2 configurations surpass CUTLASS INT1 \Msb{most of the time} despite its bit-width advantage. This superior performance can be attributed to our matrix-splitting strategy that minimizes data redundancy and our efficient memory management scheme.}

% Figure \ref{fig:llm MatMul} shows the comparison of our work with other approaches, evaluating all sizes of MatMuls in Llama-7B, using TOPS as the metric. When compared to APNN-TC, the performance improvement for smaller matrix parameters (such as 1k/1k/128) is nearly identical, but these small matrix operations do not constitute the primary part of LLM inference. For larger matrices (such as 1k/10.75k/4k), our design accelerates \msb{MatMul} more than 10 times compared to APNN-TC. This is because the design of APNN-TC is not optimized for large matrices; its allocation of shared memory and scheduling of GPU threads is more suitable for small \msb{MatMul} computations. In contrast, our design is tailored for LLMs and has been optimized specifically for large matrix computations.

\begin{figure} [t] %注意，这里设置是关键
	\centering
	\includegraphics[width=\linewidth,scale=1.00]{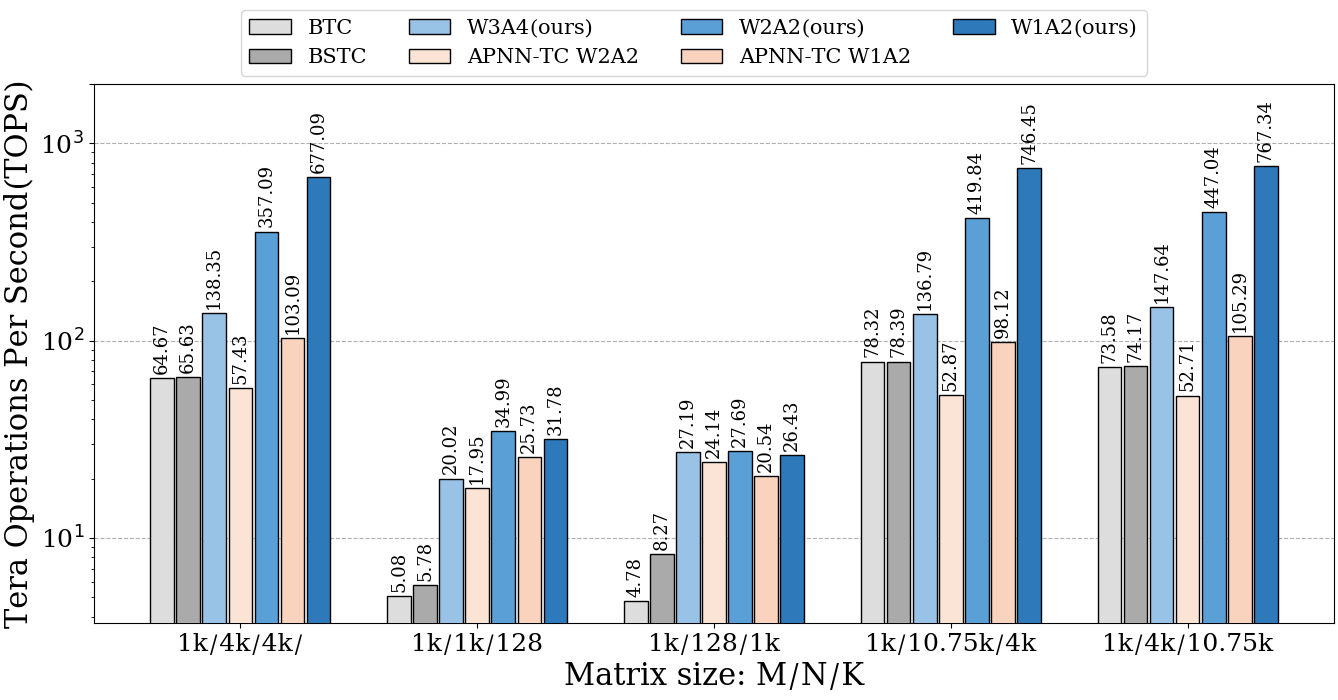}
	%[]里面的参数自己可根据需要调整
	% \caption{Comparison of our work with other approaches for MatMuls from large models.}
        \caption{\shk{Comparison of throughput between our work and other methods for typical MatMuls from LLMs.}}
	\label{fig:llm MatMul}
\end{figure}

\fc{Fig. \ref{fig:llm MatMul} illustrates the performance comparison across all MatMul sizes in Llama-7B, using TOPS as the metric. When compared to APNN-TC, our design shows similar performance for smaller matrices (e.g., 1k$\times$1k$\times$128). \Msb{However, for larger matrices more common in LLMs (e.g., 1k$\times$10.75k$\times$4k), our approach achieves over 6.7$\times$ speedup compared to APNN-TC.} This disparity is due to our design being optimized specifically for the large matrix computations prevalent in LLMs, whereas APNN-TC's shared memory allocation and GPU thread scheduling are more suitable for smaller MatMul operations.}
\fc{In summary, our MatMul kernel design demonstrates significant performance advantages in LLM-specific scenarios, particularly for larger, non-square matrices that dominate LLM computations. This makes our approach particularly well-suited for accelerating LLM inference tasks.}

\subsection{Arbitrary Precision LLM Evaluation}

\fc{Efficient inference of large language models (LLMs) is crucial for their practical deployment. We evaluate our arbitrary precision MatMul design in ultra-low bit quantized LLMs to demonstrate its effectiveness in accelerating real-world LLM inference.}

% We replaced the \msb{MatMul} operations in the LLMs with the arbitrary precision \msb{MatMul} kernel designed in this paper and compared the model inference speeds to evaluate the \fc{end-to-end speedup} of our design on ultra-low bit quantized LLMs. 
% For the evaluation, we selected three LLMs that are widely studied for quantization: Llama2-7B, OPT-6.7B, and BLOOM-7B. 
% Additionally, we set another comparison item, which is the geometric mean of the run times of the three groups of models. 
% The comparison settings include: half-precision (FP16) models running directly on PyTorch, QLoRA~\cite{qlora} (restoring 4-bit quantized weights to their original high precision before model inference.), GPTQ~\cite{gptq} (weights quantized to 4-bit, 3-bit, and 2-bit, using CUTLASS INT4 as the kernel for matrix computation.), and OneBit~\cite{onebit} (weights quantized to 1-bit, using CUTLASS INT1 as the kernel for matrix computation.) Our design configurations are: W1A1 with OneBit models, W2A2 with 2-bit quantized GPTQ models, W4A4 with 4-bit quantized GPTQ models.

\fc{We focus on three widely studied LLMs: Llama2-7B, OPT-6.7B, and BLOOM-7B. By replacing their standard MatMul operations with our arbitrary precision MatMul kernel, we seek to quantify the performance gains across different model architectures. Our comparison encompasses several state-of-the-art methods: half-precision (FP16) models running on PyTorch, QLoRA~\cite{qlora}, GPTQ~\cite{gptq}, and OneBit~\cite{onebit}. To ensure a fair comparison, we align our design configurations with these methods: W1A1 for OneBit models, W2A2 for 2-bit quantized GPTQ models, and W4A4 for 4-bit quantized GPTQ models.}

\begin{figure} [t] %注意，这里设置是关键
	\centering
	\includegraphics[width=\linewidth,scale=1.00]{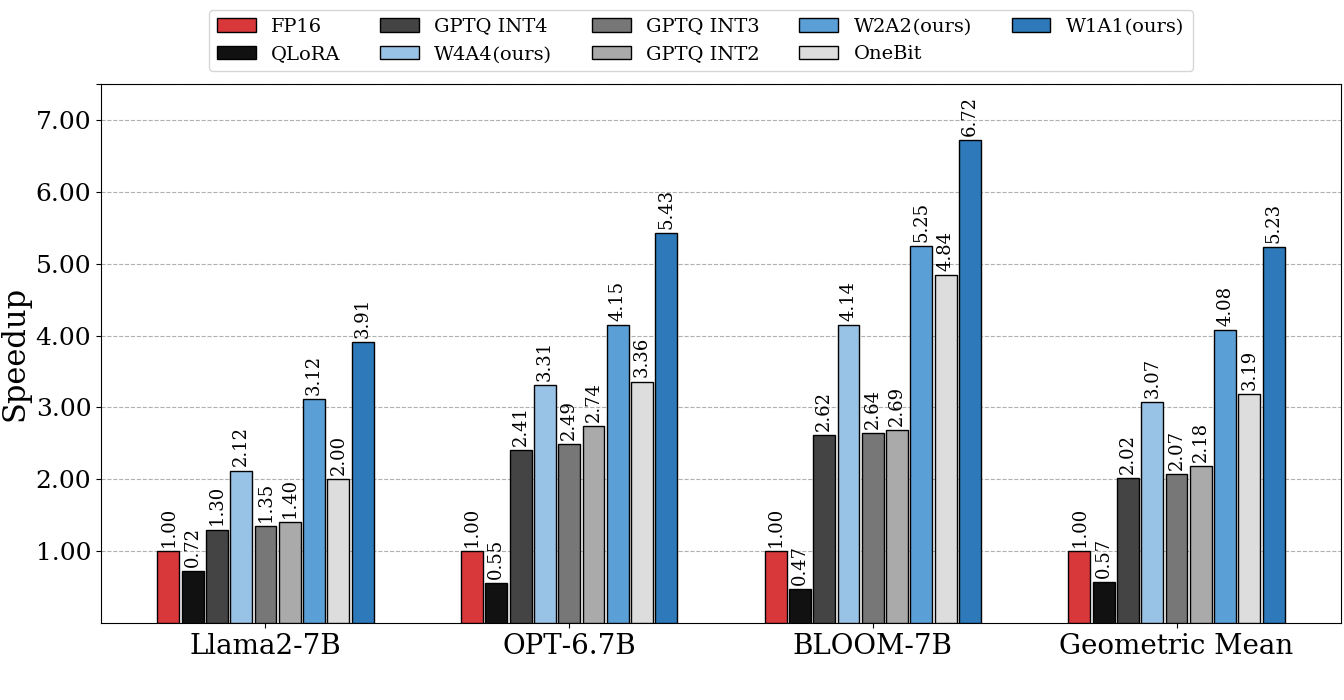}
	%[]里面的参数自己可根据需要调整
	% \caption{Comparison of inference speedup for different quantization methods in three LLMs (using FP16 as the baseline, with an additional comparison being the geometric mean of the speedup).}
        \caption{\shk{Inference speedup through different quantization frameworks for various LLMs (using FP16 as the baseline).}}
	\label{fig:llm inference}
\end{figure}

% Fig. \ref{fig:llm inference} shows that our work achieves better acceleration effects than CUTLASS under the same bit-width configuration, with up to nearly 2× speedup. Compared to PyTorch's FP16, it achieves approximately 6× speedup. Firstly, for QLoRA, its design aims to reduce memory usage rather than enhance inference speed. The precision restoration step actually makes its model inference slower. Next, for GPTQ with 2-4 bit quantization, although the quantization bit-widths are different, they are limited by the precision support of GPU \msb{TC}s. During matrix computation, some efficiency has to be sacrificed, using CUTLASS 4-bit as the kernel. The advantage of lower bit-width can only be reflected in some data transfer processes, resulting in less-than-ideal performance improvement. Finally, compared to OneBit, which also uses W1A1 operations, our \msb{MatMul} scheme is more efficient in memory scheduling, resulting in 1.2-2× faster model inference speed compared to CUTLASS.

\fc{Fig. \ref{fig:llm inference} presents the inference speedup for various quantization methods across the three LLMs, using FP16 as the baseline. The results reveal that our work achieves superior acceleration compared to CUTLASS under equivalent bit-width configurations, demonstrating up to nearly 2$\times$ speedup. Moreover, when compared to PyTorch's FP16 implementation, our approach exhibits an impressive 6$\times$ speedup.
Our analysis indicates that while QLoRA excels in reducing memory usage, its inference speed is compromised due to the necessary precision restoration step. GPTQ, despite offering 2-4 bit quantization, faces limitations imposed by GPU TC precision support. Its reliance on CUTLASS 4-bit as the kernel leads to efficiency trade-offs and suboptimal performance improvements. In contrast, our MatMul scheme demonstrates a 1.2-2$\times$ faster model inference speed compared to OneBit (W1A1), attributable to our more efficient memory scheduling.}

% Overall, both kernel-level and model-level evaluations demonstrate the flexibility in precision configuration and the effectiveness of memory management in our design. 

% In terms of \msb{MatMul}, compared to Nvidia's MatMul acceleration design, CUTLASS, our approach achieves a 5.45× speedup on square \msb{MatMul}. Additionally, compared to APNN, which also supports arbitrary precision \msb{MatMul}, our design is more suitable for large matrix computations, achieving a 43× speedup.

% For ultra-low bit quantized model inference, our design offers more flexible precision configurations, surpassing the limited precision support of GPUs and enabling 2-bit and 3-bit quantization to achieve more ideal inference results. This results in nearly a 2× speedup compared to using CUTLASS as the kernel. Furthermore, our efficient memory scheduling design allows for over a 1.2× speedup even at the same computational precision (e.g., CUTLASS INT1 and W1A1).
\section{CONCLUSIONS}
% In this paper, to accelerate the inference of ultra-low bit quantized large language models (LLMs) with arbitrary precision on GPU, we first introduce a data format called bipolar-INT, which is suitable for symmetric quantization and conducive to parallel computation. We then propose an INT \msb{MatMul} scheme applicable to arbitrary precision and design an efficient memory management system tailored for this scheme. Finally, we evaluate the performance of the arbitrary precision \msb{MatMul} and the end-to-end inference speed of LLMs. In terms of \msb{MatMul}, our implementation achieves a 5.45$\times$ speedup compared to Nvidia's CUTLASS and up to a 43$\times$ speedup compared to other works when dealing with larger matrix parameters. For LLM inference, our design showed a 3.9-6.7$\times$ speedup over FP16 models, and a 1.2-2$\times$ speedup compared to ultra-low bit quantized models using cutlass as the kernel.

\fc{In this paper, we present a comprehensive scheme to accelerate inference of ultra-low bit quantized large language models (LLMs) on GPU.}
\fc{We first introduce a data format called bipolar-INT, which is suitable for symmetric quantization and conducive to parallel computation.}
On top of bipolar-INT, we propose an INT \msb{MatMul} method applicable to arbitrary precision and design an efficient memory management system.
\fc{Our evaluations demonstrate significant performance gains in both MatMul operations and LLM inference. \Msb{For MatMul, our implementation achieves a 2.4$\times$ speedup compared to Nvidia's CUTLASS.} In LLM inference, our approach yields 3.9-6.7$\times$ speedup over FP16 models and 1.2-2$\times$ speedup compared to ultra-low bit quantized models using CUTLASS as the kernel.}
% \fc{These results highlight the effectiveness of our approach in accelerating ultra-low bit quantized LLMs, offering substantial improvements in computational efficiency and inference speed.}

\begin{acks}
This work was supported in part by the National Key R\&D Program of China under Grant 2022YFB4400600 and in part by the Postgraduate Research \& Practice Innovation Program of Jiangsu Province under Grant KYCX24\_0149.
\end{acks}

\bibliographystyle{ACM-Reference-Format}
\bibliography{sample}

\end{document}